\pdfoutput=1
\documentclass[11pt]{article}

\usepackage[]{acl2023}

\usepackage{times}
\usepackage{latexsym}

\usepackage[T1]{fontenc}
\usepackage[utf8]{inputenc}

\usepackage{microtype}

\usepackage{hyperref}
\usepackage{url}

\usepackage{epsfig}
\usepackage{amsmath}
\usepackage{amssymb}
\usepackage{microtype}
\usepackage{xspace}
\usepackage{subfig}
\usepackage{graphicx,wrapfig}
\usepackage{booktabs}
\usepackage{makecell}
\usepackage{framed}
\usepackage{multirow}
\usepackage[linesnumbered,algoruled,boxed,noend]{algorithm2e}
\usepackage{enumitem}
\usepackage{bm}

\usepackage{pifont}

\newcolumntype{L}[1]{>{\raggedright\let\newline\\\arraybackslash\hspace{0pt}}m{#1}}
\newcolumntype{C}[1]{>{\centering\let\newline\\\arraybackslash\hspace{0pt}}m{#1}}
\newcolumntype{R}[1]{>{\raggedleft\let\newline\\\arraybackslash\hspace{0pt}}m{#1}}

\newcommand{\para}[1]{\noindent\textbf{#1}}
\newcommand{\methodn}{\textsc{GRILL}}%
\newcommand{\method}{\methodn\xspace}

\usepackage{titlesec} % title spacing
\titlespacing*{\section}{1pt}{6pt}{2pt}
\titlespacing*{\subsection}{1pt}{6pt}{2pt}
\titlespacing*{\subsubsection}{1pt}{1pt}{1pt}

\title{GRILL: Grounded Vision-language Pre-training\\via Aligning Text and Image Regions}
\author{
    Woojeong Jin$^1$\thanks{\xspace\xspace Work was mainly done while interning at Microsoft Research.}, Subhabrata Mukherjee$^2$, Yu Cheng$^2$, Yelong Shen$^2$,\\ \textbf{Weizhu Chen$^2$, Ahmed Hassan Awadallah$^2$, Damien Jose$^2$, Xiang Ren$^1$}\\
    $^1$University of Southern California \quad $^2$Microsoft Corporation \\
    {\small \texttt{\{woojeong.jin,xiangren\}@usc.edu}}\\
    {\small \texttt{\{submukhe,yu.cheng,yelong.shen,wzchen,hassanam,dajose\}@microsoft.com}}\\
}

\begin{document}
\maketitle
\begin{abstract}

% \xiang{Our articulation should highlight and focus on the zero-shot/few-shot generalization ability; rather than pitching a general VL pretraining narrative.}

Generalization to unseen tasks is an important ability for few-shot learners to achieve better zero-/few-shot performance on diverse tasks.
However, such generalization to vision-language tasks including grounding and generation tasks has been under-explored; existing few-shot VL models struggle to handle tasks that involve object grounding and multiple images such as visual commonsense reasoning~\citep{zellers2019recognition} or NLVR2~\citep{suhr2018corpus}.
In this paper, we introduce \method, \textbf{GR}ounded v\textbf{I}sion \textbf{L}anguage a\textbf{L}igning, a novel VL model that can be generalized to diverse tasks including visual question answering, captioning, and grounding tasks with no or very few training instances.
Specifically, \method learns object grounding and localization by exploiting object-text alignments, which enables it to transfer to grounding tasks in a zero-/few-shot fashion. 
We evaluate our model on various zero-/few-shot VL tasks and show that it consistently surpasses the state-of-the-art few-shot methods.

\end{abstract}

\section{Introduction}
\textit{Generalization to unseen tasks} has been explored and investigated on zero-/few-shot NLP tasks by performing multi-task learning with task-specific prompts~\citep{sanh2021multitask} or pre-training huge language models on a massive dataset and using a few examples as demonstrations for generalization~\citep{brown2020language}.
Similarly, few-shot vision-language (VL) learning methods aim to leverage the pre-trained language models and their powerful generalization abilities to adapt to VL domains and learn new tasks from zero or a few examples~\citep{tsimpoukelli2021multimodal,radford2021learning,jin2021good,alayrac2022flamingo}.

% \xiang{the pitch for the 1st para would be the cross-task generalization to a broader set of VL (grounded) tasks is under-explored and is the focus of your paper.}
% Language transformers, such as GPT-3~\citep{brown2020language} and T0~\citep{sanh2021multitask}, have achieved state-of-the-art results on various natural language understanding tasks without any task-specific fine-tuning.
% They only rely on large-scale pre-training on diverse text corpora and a natural language query as the input or prompt to perform different tasks. 
% Similarly, few-shot learning methods aim to leverage the pre-trained language models and their powerful generalization abilities to adapt to vision-language (VL) domains and learn new tasks from zero or a few examples~\citep{tsimpoukelli2021multimodal,radford2021learning,jin2021good,alayrac2022flamingo}. 
% These methods have shown promising results on tasks that require understanding of the semantic and syntactic relations between images and natural language, such as, visual question answering (VQA)~\citep{goyal2017making,marino2019ok,hudson2019gqa} where the model has to answer a natural language question about an image; and image captioning~\citep{lin2014microsoft,agrawal2019nocaps,young2014image}, where the model has to generate a natural language description of an image.

\begin{figure}[tb!]
    \centering
    {\includegraphics[width=\linewidth]{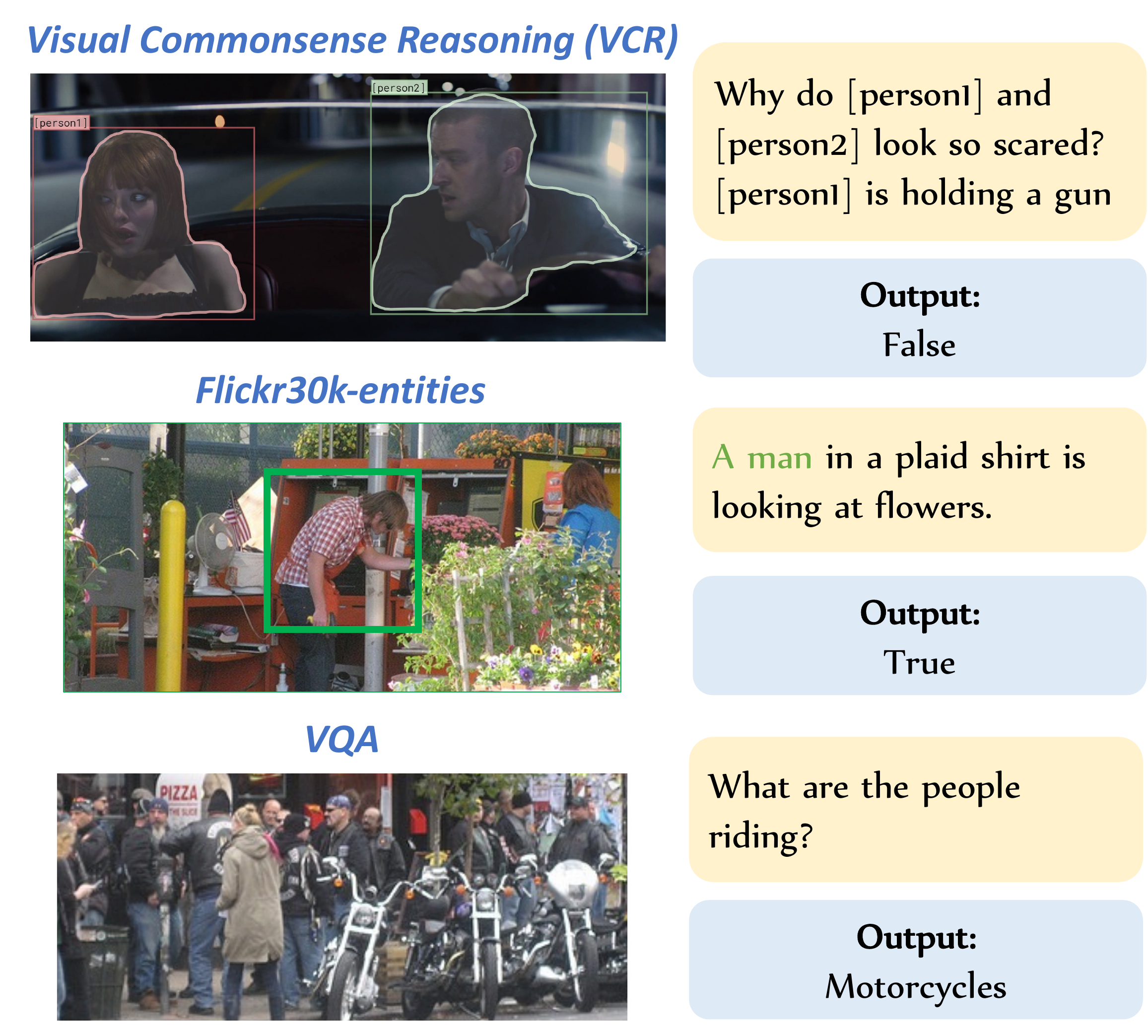}} 
    \caption{\textbf{Examples of vision-language tasks.} Vision-language tasks have different task formats, which makes challenging to generalize in a zero-/few-shot way. In this work, we study generalization of few-shot methods and propose \method that  can generalize to diverse VL tasks without introducing task-specific special representations or pre-trained object detectors.
    % including grounding tasks in a zero-/few-shot manner. \wj{[person 1]}
    }
    \label{fig:intro_example}
\end{figure}

While the few-shot learners can overcome the challenges of supervised learning and avoid the need for task-specific fine-tuning, existing few-shot VL learners 
suffer from \textit{limited generalization to unseen tasks such as grounding tasks}
% {\em do not address the challenge of grounding tasks} 
that require not only understanding the image and the language, but also locating and identifying relevant regions or objects in images, such as visual commonsense reasoning (VCR)~\citep{zellers2019recognition} or Flickr30k-entities~\citep{plummer2015flickr30k}.
% where the model has to reason about the actions, intentions, and emotions of agents in the image; or Flickr30k-entities~\citep{plummer2015flickr30k}, where the model has to align the mentions of entities in the captions with their corresponding regions in the image. 
% These tasks are essential for VL models to achieve human-like reasoning and understanding. 
% However 
Existing few-shot VL methods exhibit great performance on visual question answering and captioning tasks~\cite{alayrac2022flamingo,tsimpoukelli2021multimodal,jin2021good}, but they lack the skills to generalize to grounding tasks as they {\em do not explicitly model the spatial and visual information of the regions or objects}.
On the other hand, existing fine-tuning methods rely on special representations for representing regions or objects, such as special tokens that mark the regions or objects in the captions and the images~\citep{cho2021unifying}, and object features extracted from a pre-trained object detector~\citep{su2019vl,chen2019uniter}. 
These methods achieve good results with fine-tuning, but they are not compatible with zero-/few-shot generalization, due to the different designs of object representation for each task and the dependence on external object detectors that may not cover all the relevant concepts.

% our task format and input

% our method

In this paper, we introduce \method, \textbf{GR}ounded v\textbf{I}sion \textbf{L}anguage a\textbf{L}igning, a new VL model that can be generalized to diverse tasks including visual question answering, captioning, and grounding tasks in a zero-/few-shot fashion.
We address the challenge of few-shot generalization to unseen tasks by a) learning object grounding and localization in pre-training, b) representing visual concepts (e.g., regions and images) with versatile image patches, and c) unifying the tasks into text generation.
% that can learn object grounding and localization during pre-training and generalize to a wide range of VL tasks including visual question answering, captioning, and grounding tasks in a zero-/few-shot fashion.
Specifically, our model is a generative sequence-to-sequence transformer model~\citep{vaswani2017attention} with a vision transformer (ViT)~\citep{dosovitskiy2020image,liu2021swin} to process images with patch embeddings, where each patch represents a fixed-size region of the image. 
We represent a visual concept (object or region) that corresponds to a group of patches by aggregating information across the patches. 
This enables our model to generate better representations for \textit{any kind of regions or images}.
% without relying on pre-trained object detectors, which may be noisy, incomplete, or biased. 
% Thus, our model can handle diverse VL tasks that involve region/object grounding in a zero-/few-shot manner, by simply taking an image and a natural language query as the input and producing a natural language answer as the output.
We construct our pre-training dataset from MS-COCO~\citep{lin2014microsoft,chen2015microsoft} and Visual Genome~\citep{krishna2017visual}, where each caption contains images or bounding boxes within them, which provide rich and diverse information for the model to learn object grounding and localization.
Given the dataset, we pre-train our model with prefix language modeling (PrefixLM) and masked language modeling (MaskedLM) objectives, which encourage the model to generate natural language from images and fill in the missing words in captions, respectively; and a discriminative objective, which encourages the model to distinguish whether the paired image-captions are correct or not.

% our experiment
We test our \method on $7$ zero-/few-shot vision-language tasks including Visual Commonsense Reasoning (VCR)~\citep{zellers2019recognition}, RefCOCOg~\citep{mao2016generation},  Flickr30k-entities~\citep{plummer2015flickr30k}, NLVR2~\citep{suhr2018corpus}, SNLI-VE~\citep{xie2019visual}, visual question answering~\citep{goyal2017making}, and Flickr30k captioning~\citep{young2014image}. 
% Our \method consistently surpasses the state-of-the-art few-shot competitors.
We observe that our model demonstrates better zero-/few-shot generalization on diverse tasks compared to baselines. 
We also find that our pre-training objectives and pre-training datasets are vital for better zero-/few-shot performance.
% Furthermore, we observe that our grounded pre-training data is important for the improvement.
% Our \method shows \wj{results}.
% Furthermore, we observe that (1) ..., (2) ..., and (3) ... \wj{update}

\begin{figure}[tb!]
    \centering
    {\includegraphics[width=\linewidth]{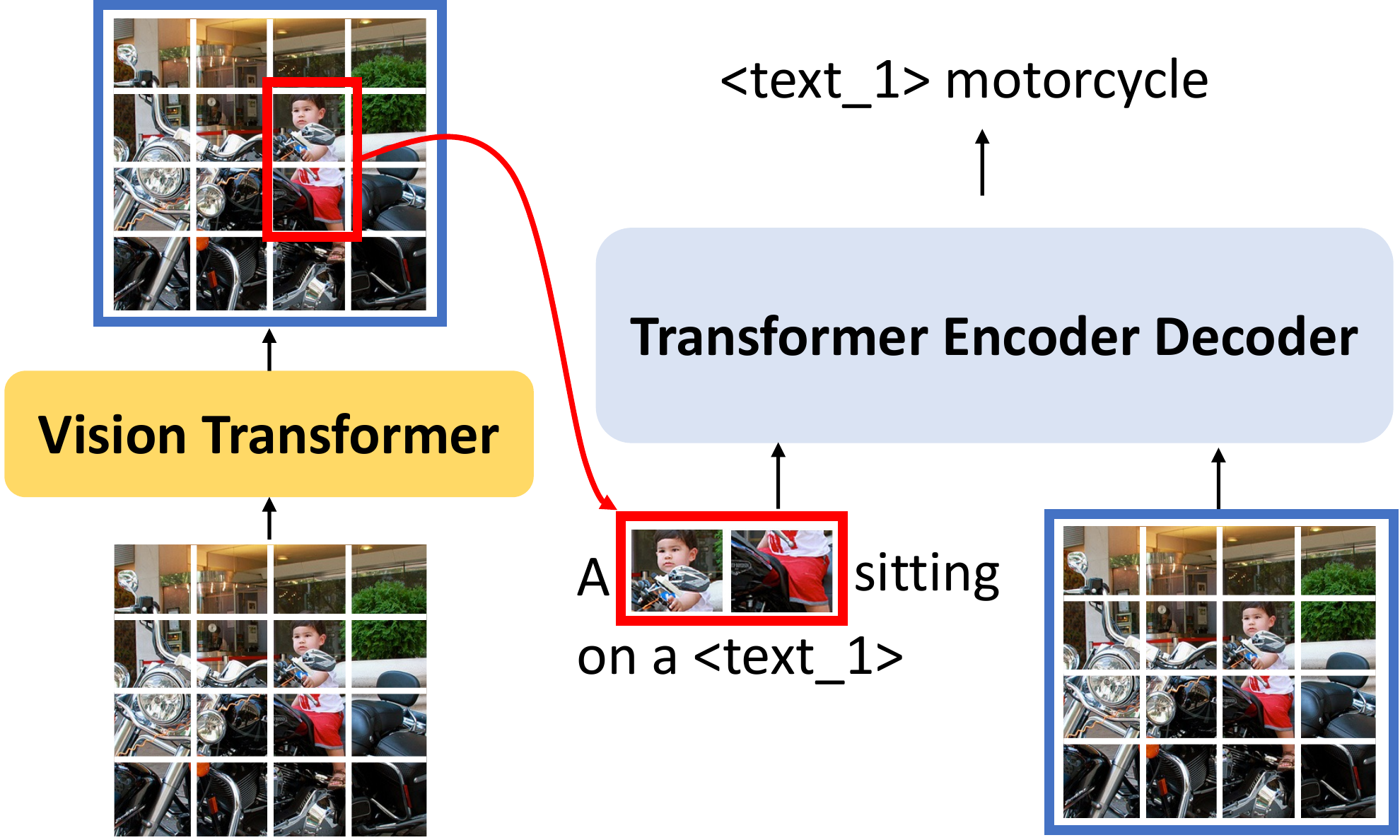}} 
    \caption{\textbf{Illustration of \method.} Our model is a sequence-to-sequence transformer that uses a vision transformer (ViT)~\citep{dosovitskiy2020image,liu2021swin} to process images with patch embeddings, where each patch represents a fixed-size region of the image. We replace the referring words with the corresponding visual patches.
    % \xiang{Convert to a single page figure, or compressing the blank space.}
    }
    \label{fig:illust}
\end{figure}

% \section{Problem Setup and Challenges}
\section{Generalization to Diverse Vision-language Tasks}

% We introduce our analysis setup: problem formulation, analysis questions, downstream tasks and datasets, evaluation metrics, and baselines. 
% \xiang{This paper is more like a method/model paper, than an analysis paper. For the problem formulation part, you can describe the goal of ``zero-shot generalization to a variety of VL tasks", including grounded VL tasks; and give formal notations and definition in terms of input/output and evaluation.}
% \xiang{Sec 2 organization is a bit confusing now. You probably want to separate out the background stuffs into a separate section, or merge into Sec 3}
% \xiang{To make the paper more exciting to read, maybe you should pitch the problem framing to be a novel contribution. So the Sec 2 can be named as the new challenge and try to articulate it to be more exciting.
% ghen Sec 3 is the proposed solution for this new challenge, plus giving background on the existing work which can be applied to this problem, yet limited.}
% \xiang{Also, good to include another 2-3 illustrative figure, to illustrate all the objectives. and a figure for the problem setting}

% \xiang{Needs major revision to tweak the framing and align with key msgs in Intro.}
% Given one or more images and a caption with bounding boxes from the images, can VL models understand the caption and the images?

Various VL tasks require phrase and object grounding and their task formats are different, which makes few-shot models challenging to generalize.
In this work, we introduce a model that can generalize to VL tasks including grounding with \textit{no} or \textit{a few} labeled examples.
We first introduce the background, formal problem definition, and challenges.

% Can we jointly leverage images and captions with bounding boxes for learning to ground in a zero-/few-shot fashion?
% In this work, we present an interesting challenge that requires understanding of the text with multiple images that may refer to objects, regions, or the entire image in a zero-/few-shot fashion.
% Our problem setup is unique and challenging since there are diverse formats of VL tasks, and models are required to understand and solve the tasks with \textit{no} or \textit{a few} training examples.
% We first introduce the background, formal problem definition, and challenges.

% The goal of this work is to pre-train a VL model $\mathcal{L}$, and evaluate the model on VL tasks that require grounding between text and regions or images in a zero-/few-shot manner. 
% We introduce the definition of grounding and our problem formulation.

\subsection{Background: Visual Grounding}
\label{sec:method:background}
Visual grounding refers to the ability to link linguistic concepts (sentences, phrases, or words) to visual concepts (images and regions)~\citep{chandu2021grounding}.
Here we consider two types of visual grounding: image grounding and object grounding.
% \para{Image grounding.}
%Image grounding has been widely used in many pieces of literature without a clear definition.

\textit{Image grounding} refers to the linking of textual concepts to image concepts~\citep{chandu2021grounding}.
% This definition subsumes any type of visual grounding and is quite broad in scope. % and it is a too comprehensive definition.
In this work, we consider image grounding as linking any type of text including sentences, phrases, and words to an entire image (e.g., image captioning, and image retrieval).
% image captioning and image retrieval tasks are examples of image grounding in our definition.
% \para{Object grounding.}
Given an image and a corresponding caption, \textit{object grounding} aims to localize objects  in the image as mentioned by a noun phrase in the caption (or the entire caption sentence). 
% \xiang{Add a citation for definition of this term.}
Such object grounding occurs at word, phrase, and sentence levels in the language modality.
% Object grounding is also called phrase grounding or localization.
Many VL tasks require object grounding implicitly or explicitly and we consider tasks that explicitly require localization as object grounding tasks such as  referring expression comprehension (RefCOCOg~\citep{mao2016generation}), phrase grounding (Flickr30k-entities~\citep{plummer2015flickr30k}), and visual commonsense reasoning~\citep{zellers2019recognition}.
% are examples of localization tasks.

\subsection{Problem Formulation}
In this work, we re-formulate the widely used pre-training task for image-caption datasets such that each caption may have one or more images including bounding boxes or regions in itself as a part of the text, denoted by $(T,\{V_j\}^N)$, in addition to the associated images. 
Note that some captions may not have images in themselves, $N=0$.
We refer to learning on the captions with images \textit{grounded} learning. 
For pre-training, a VL model is pre-trained on image-caption datasets where captions include images or bounding boxes. 
% We aim to learn grounding and localization through the pre-training.
For zero-shot tasks, the pre-trained model $\mathcal{L}$ cannot access training data $\mathcal{D}_{train}$ and validation data $\mathcal{D}_{val}$. 
We directly evaluate the model on the test data $\mathcal{D}_{test}$.
For few-shot tasks, the model has access to $K$ instances of training data for fine-tuning.
For hyper-parameter tuning and model selection,
we assume validation data $\mathcal{D}_{val}$ which has an equal number of instances to $\mathcal{D}_{train}$ to simulate a real-world low-resource environment and compose the validation data from training data.
The sizes of $\mathcal{D}_{train}$ and $\mathcal{D}_{val}$ are 32 in our study.

\paragraph{Challenges}
Our goal is to pre-train a VL model that seamlessly transfers to various tasks not limited to visual question answering and captioning in a zero-shot or few-shot manner.
% visual question answering and captioning, .
% This paper focuses on grounding tasks that require image grounding or object grounding. 
Different tasks, especially grounding tasks, have different task (input and output) formats as in Fig.~\ref{fig:intro_example}, and thus the main challenge of this work is to generalize the zero-/few-shot ability to diverse tasks.
% We 
% unifies the formats and tests zero-/few-shot ability on the different tasks.
Existing works on grounding tasks introduce special representations to depict regions such as special tokens~\citep{cho2021unifying} or object representations by an object detector~\citep{su2019vl,chen2019uniter}. 
While these works perform well on grounding tasks via expensive fine-tuning on labeled data, they have to design different object representations for different task formats. This makes it difficult to generalize to new tasks in a zero-shot fashion.
For example, the object representations from an object detector are difficult to transfer to a task that refers to multiple images such as NLVR2~\citep{suhr2018corpus}. 
% \wj{maybe figure?}
In this work, we tackle these challenges by introducing patch embeddings to represent objects, regions, and images; learning object grounding and localization in pre-training, and unifying all the tasks into text generation.

% \para{Challenge.}

% \section{Method}
\section{Pre-training for Better Task Generalization}

% \xiang{Have a paragraph to overview the key components of your approach.}

In this section, we introduce \method, a few-shot VL model for jointly learning contextualized representations from vision and language tasks.
We first present an overview of \method (\S\ref{sec:method:overview}), our model architecture (\S\ref{sec:method:architecture}), pre-training objectives (\S\ref{sec:method:obj}), and pre-training data (\S\ref{sec:method:data}) in this section.

% \wj{what is grounded sequence}

\subsection{Overview}
\label{sec:method:overview}

% \xiang{just called it “hybrid sequence”,  as it is short for text-img region hybrid sequence.}

We propose \method, a VL model that can learn object grounding and localization in pre-training and generalize to a wide range of VL tasks in a zero-/few-shot fashion.
Our model is a sequence-to-sequence transformer ~\citep{vaswani2017attention} and takes a \textit{hybrid sequence}, denoted by $(I,T,\{V_j\}^N)$, consisting of text $T$, an image $I$ and visual concepts or regions $\{V_j\}^N$ as input and the output is a text sequence. 
We represent an input image with image patches by vision transformer~\citep{dosovitskiy2020image,liu2021swin} and represent a region that corresponds to a set of patches by aggregating information among the patches (\S\ref{sec:method:architecture}).
We illustrate our model in Fig.~\ref{fig:illust}.
Given sequences with paired text outputs, we pre-train our model with prefix language modeling, masked language modeling, and a discriminative objective (\S\ref{sec:method:obj}).
Then we discuss how we create the hybrid sequences from image-caption datasets (\S\ref{sec:method:data}).
% \wj{we tackle the challenge by introducing patch embeddings to represent objects, regions, and an image itself in the pre-training.}

% \subsection{Grounded Sequence}

% \subsection{encoding image/region/text}

\subsection{Model Architecture}
\label{sec:method:architecture}
For unified text generation, we adopt a transformer encoder-decoder architecture~\citep{vaswani2017attention}, which takes a text sequence as an input and generates another text sequence as an output.
To encode images and regions for vision-language tasks, we adopt a vision transformer~\citep{dosovitskiy2020image,liu2021swin} as our image encoder; it splits an input image with a sequence of image patches.
Specifically, it first splits an image into non-overlapping patches and linearly embeds all patches, and these patches are passed to the transformer encoder layers, yielding $\{v_1, ..., v_m \}$. For an image of resolution of $224 \times 224$ and patch size of $32 \times 32$, we have $m = 49$.
We assume that $v_i$ encodes the information of the corresponding patch $p_i$. 
The image patches are versatile in that they can represent any type of images or regions; we represent a visual concept (object or region) $V_j$ that corresponds to a set of patches by aggregating information among the patches, and these patches are additionally passed to the transformer encoder layer.
% to encode visual and text inputs and generate target text.
% In addition to the encoder-decoder architecture, we introduce an image encoder to
% We treat an input image with a sequence of image patches by a vision transformer~\citep{dosovitskiy2020image,liu2021swin} to better represent any type of images and regions. 
We adopt Swin transformer (Swin-B)~\citep{liu2021swin} as our vision transformer.
% Details of the model architecture are described in Sec.~\ref{append:arch} of the appendix.

% it first splits an image into non-overlapping patches and linearly embeds all patches, and these patches are passed to the transformer encoder layers, yielding $\{v_1, ..., v_m \}$. For an image of resolution of $224 \times 224$ and patch size of $32 \times 32$, we have $m = 49$.
% We assume that $v_i$ encodes the information of the corresponding patch $p_i$. 
% Therefore, we represent a visual concept (object or region) $V_j$ that corresponds to a set of patches by aggregating information among the patches as shown in Fig.~\ref{fig:illust}.
% In addition, the entire patch representations are fed into the encoder by appending them to the text to encode the whole image.
% Then the transformer decoder outputs the target text.

% The model is not task-specific, so it is a good option for zero/few-shot settings.
% \wj{size of transformers. How the swin is used?}

% \para{Representing bounding boxes.}
% Representing bounding boxes is 
% \wj{add how to get bounding boxes from patches}

\begin{figure}[tb!]
    \centering
    {\includegraphics[width=0.99\linewidth]{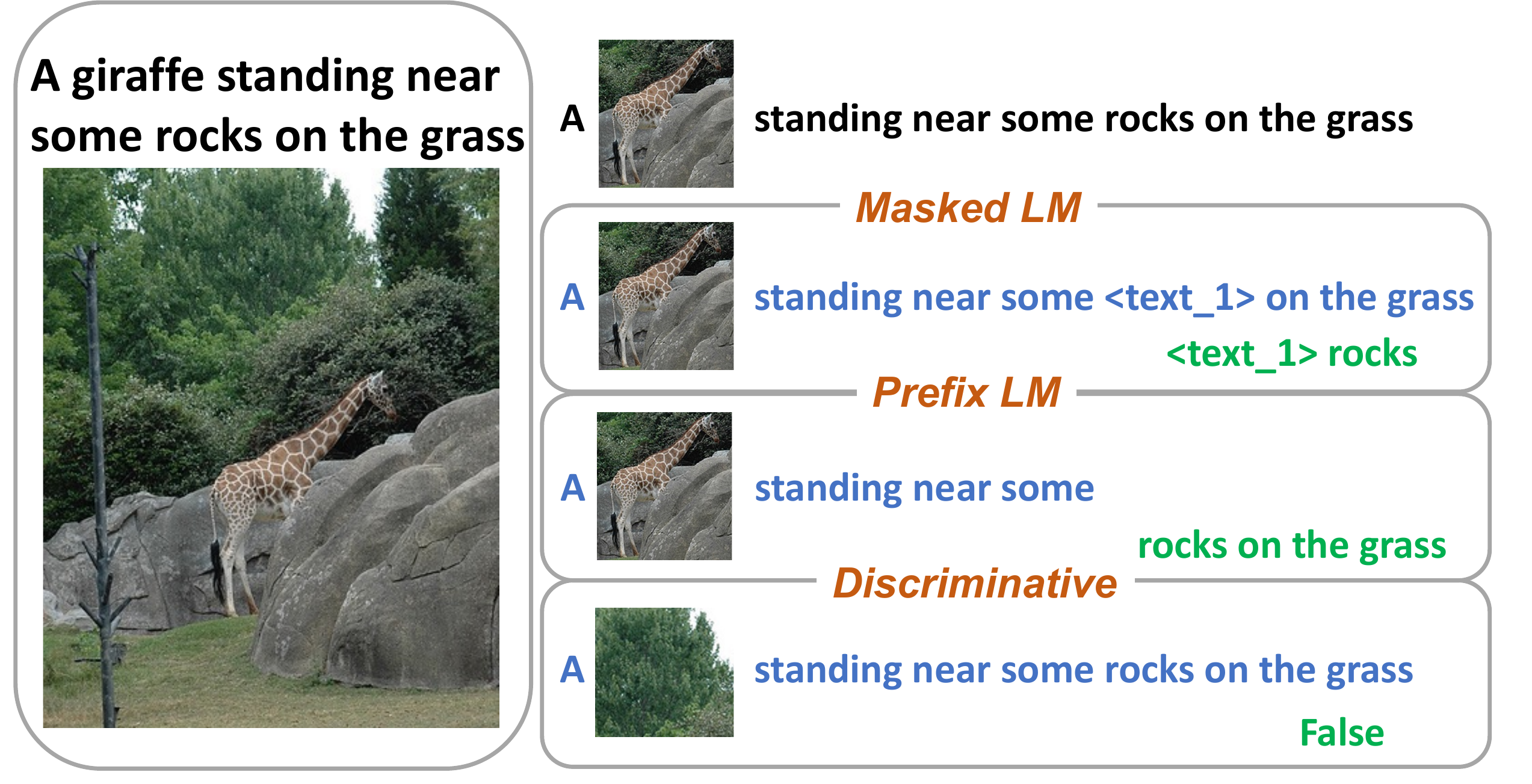}} 
    \caption{\textbf{Pre-training objectives.} We illustrate our pre-training objectives. We include masked language modeling, prefix language modeling, and the discriminative objective as our pre-training objectives. Given an image-caption pair, we create proper inputs for each objective. Text in green color is the target text of each objective.
    }
    \label{fig:data_example}
\end{figure}

\subsection{Pre-training Objectives}
\label{sec:method:obj}
We pre-train our model with prefix language modeling (PrefixLM), masked language modeling (MaskedLM) following \citet{jin2021good}, and a discriminative objective.
Many VL tasks are classification tasks that require choosing one of the options.
To deal with the classification tasks, we additionally adopt the discriminative objective, which is to classify whether the given sequence is correct or not.
% We aim to learn grounding and localization through pre-training.
% Each caption may have region representations in itself, so we carefully design the objectives.
% Note that our model takes hybrid an image and text with regions as inputs and generates target text.
% In addition to the hybrid sequences, we also include original text and images as our pre-training data.
Fig.~\ref{fig:data_example} illustrates the pre-training objectives.

\para{Prefix language modeling.}
We include prefix language modeling (PrefixLM) following \citep{raffel2019exploring,jin2021good}.
The objective randomly splits the text with regions input into two separate sequences. 
The first part may contain regions and is used as an input with an image to the encoder, and the second part does not contain regions and is used as target text to be generated by the decoder.
The target text is not allowed to have region representations since our model generates text only
% Given an image and a span of text with region representations, this objective randomly splits the text into two separate components; the former component with the given image is used as inputs to the encoder and the latter component is used as target text to be generated by the decoder.
% We allow the input text to include region representations while the target text is not allowed to have them since our model generates text only.

\para{Masked language modeling.}
Masked language modeling~\citep{cho2021unifying,jin2021good} is to mask out random spans with numbered sentinel tokens, e.g., \texttt{<text\_1>}, and then the masked sequence is fed into the encoder.
% We follow \citept{cho2021unifying} to do masked language modeling.
% This objective 
Then the decoder generates the masked spans as target text.
We randomly mask 15\% of input text tokens and replace them with sentinel tokens.
Note that the input sequence may include region representations in addition to a paired image and the region representations are not allowed to be masked.

\para{Discriminative objective.}
% \wj{why disc?}
The discriminative objective is important so that our model can do classification tasks where it has to determine whether the given sequence is correct or not.
Thus, we pre-train \method with the discriminative objective and the model generates target texts, ``true'' for positive pairs and ``false'' for negative pairs.
We consider an image and its captions with associated regions (if any) as positive pairs. 
With a probability of 50\%, we create the negative pairs by replacing the referring words with random region representations from the given image or randomly choosing another training caption.
% and replace one of words in the sentence with one region representation from the given image.
The negative samples let the model learn the correct bindings of referring words and corresponding regions.
% For the raw text and raw image pairs, we randomly sample another training caption to create a negative pair.

% If there is no detected regions in the image, then we replace the given caption with another caption and treat it as negative.

% With a probability of 50\%, we randomly sample another training caption to create a negative pair. 
% We replace words with random region representations from the given image and treat this as a negative pair. 
% The model then generates target texts, ``true'' for positive pairs and ``false'' for negative pairs.
% \wj{negative pair}

% \wj{Negative}

% word-object alignments

% \begin{figure}[tb!]
%     \centering
%     {\includegraphics[width=0.3\linewidth]{figures/fake.pdf}}
%     \caption{\textbf{Dataset Creation figure} 
%     }
%     \label{fig:data}
% \end{figure}

% \begin{figure}[tb!]
%     \centering
%     {\includegraphics[width=1\linewidth]{figures/obj_data.pdf}}
%     \caption{\textbf{Objective figure} Data creation
%     }
%     \label{fig:data}
% \end{figure}

\subsection{Pre-training Data}
\label{sec:method:data}
To pre-train \method, we collect image-caption data from MS COCO~\citep{lin2014microsoft,chen2015microsoft} and Visual Genome (VG)~\citep{krishna2017visual}.
% The pre-training datasets contain 6M image-text pairs and 180K distinct images.
From the image-caption pairs, we create our hybrid sequences which may have one or more region representations pre-training.
% , where each caption may have one or more region representations. 
% We also include raw text and raw images (non-hybrid sequences) as our pre-training data.
We introduce object-word alignments representing correspondence between words and objects, and use the alignments to create hybrid sequences.
% replace a word in each caption with a corresponding region representation.
% Given the object-word alignments, we replace words in a caption with a corresponding region representation so that the caption has a region representation as a substitute for the aligned word.
We create hybrid sequences in pre-training on the fly; we randomly choose $k$ object-word alignments and replace the words with the corresponding bounding boxes. 
In addition, we include region descriptions and the aligned regions as hybrid sequences from Visual Genome, and non-hybrid sequences (raw text and images) in the pre-training.
% \wj{add VG version?}
% \wj{more details of data}

\begin{figure}[tb!]
    \centering
    {\includegraphics[width=0.99\linewidth]{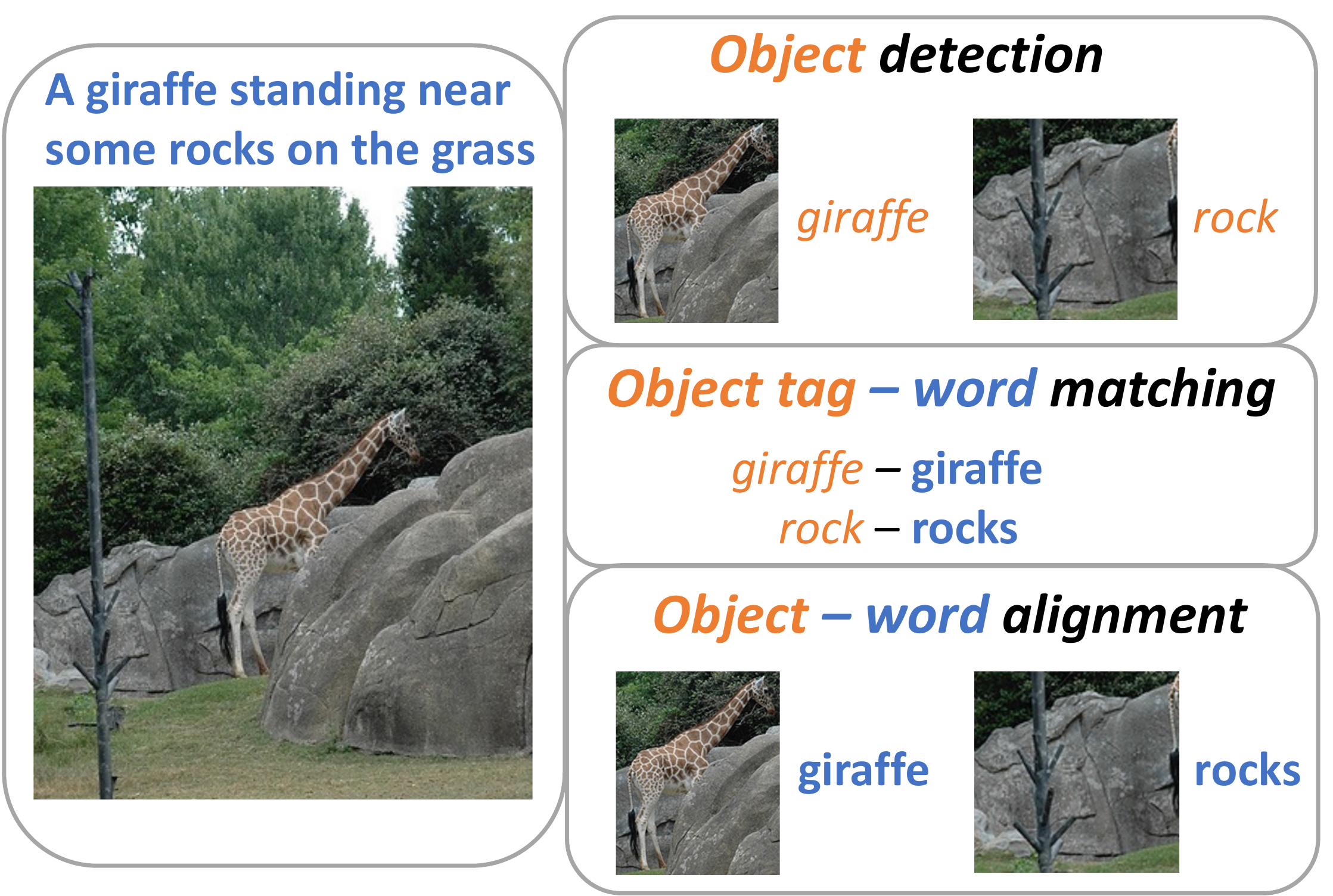}} 
    \caption{\textbf{Object-word alignments.} To create hybrid sequences, we first get object-word alignments by object detection, object tag-word matching, and object-word alignments.
    % which is to replace referring words with corresponding bounding boxes.
    }
    \label{fig:data_creation}
\end{figure}

\subsubsection{Object-word Alignments}
Given image-caption pairs, the process of getting object-word alignments consists of three steps: (1) object detection on images, (2) object tag-word matching, and (3) object-word alignments. We illustrate the process in Fig.~\ref{fig:data_creation}.
Note that we use object detection only in pre-training and do not use it on downstream tasks.

\para{Object detection.}
The first step is to detect objects and object tags from images.
We use the state-of-the-art object detector~\citep{zhang2021vinvl} to get object bounding boxes and tags, yielding $\{(V_1, l_1), ..., (V_m, l_m) \}$ where $V_i$ is a bounding box and $l_i$ is a tag for the box.
% We first detect objects in images and tags for the objects using the state-of-the-art object detector~\citep{zhang2021vinvl}, yielding $\{(V_1, l_1), ..., (V_m, l_m) \}$ where $V_i$ is a bounding box and $l_i$ is a tag for the box.
Given the set of tags $\{l_1,...,l_m \}$, we will find correspondence between the tags and words $\{w_1, ..., w_n\}$ in a caption in the next step.
% and then find correspondence between object tags and words in a caption. 

\para{Object tag-word matching.}
The second step is to find similar words $\{w_1, ..., w_n\}$ to one of tags $\{l_1,...,l_m \}$.
To find similar words, we introduce a rule-based approach as follows:
\begin{itemize}[noitemsep,topsep=0pt]
  \item Exact token matching
  \item Plural - Singular exact token matching
  \item Word vector similarity~\citep{mikolov2013distributed}
  \item WordNet Synonyms~\citep{miller1995wordnet}
\end{itemize}
If one of the rules is satisfied, then we mark them as aligned tags and words $\{(l_i, w_j)\}$.
Note that a word can be matched to multiple tags.

\para{Object-word alignments.}
In the last step, we find alignments between object bounding boxes and words $\{(V_i, w_j)\}$ given the alignments between tags and words $\{(l_i, w_j)\}$ and an object list $\{(V_1, l_1), ..., (V_m, l_m)\}$. 
% Thus, we get $\{(o_i,l_i,w_j) \}$.
We can simply find the object-word alignments since each tag is mapped to each bounding box, yielding $\{(V_i,l_i,w_j) \}$.
However, note that some object bounding boxes share the same object tag; thus the alignments can include noisy correspondences between object boxes and words.
% it is important to find the most plausible alignment.
To filter out the noisy alignments, we run CLIP~\citep{radford2021learning} over the aligned words and objects.
% To consider contextualized word representations in a caption, we use a window that includes the word we are interested in to find alignments. \wj{need to improve language here}
After this process, we obtained 1.8 object-word alignments per image-caption pair on average.

% \wj{data}
% \para{VG version}

% \subsection{Inference}

\section{Experiments}

\begin{table*}[!t]
	\centering
	\small
	\resizebox{\textwidth}{!}{
		\begin{tabular}{l|c|ccccccccccc}
            \toprule
           \multirow{2}{*}{\textbf{Method} } & \multirow{2}{*}{\textbf{Size} } &  \multicolumn{3}{c}{\textbf{VCR}} & \multicolumn{1}{c}{\textbf{RefCOCOg}} &  \multicolumn{3}{c}{\textbf{Flickr30k-entities}}  & \multicolumn{1}{c}{\textbf{NLVR2}} & \multicolumn{1}{c}{\textbf{SNLI-VE}} & \multicolumn{1}{c}{\textbf{VQAv2}} & \multicolumn{1}{c}{\textbf{Flickr30k}}\\
            \cmidrule(lr){3-5} \cmidrule(lr){6-6}     \cmidrule(lr){7-9} \cmidrule(lr){10-10}  \cmidrule(lr){11-11} \cmidrule(lr){12-12} \cmidrule(lr){13-13} 
            & & Q $\rightarrow$ A & QA $\rightarrow$ R & Q $\rightarrow$ AR & Acc & R@1 & R@5 & R@10 & Acc & Acc & Acc & CIDEr \\
            \midrule
            Random              & -     & 25.0  & 25.0  & 6.3   & 19.0  & 6.5   & 27.7  & 47.8  & 50.0  & 33.3  & 0.0   & - \\
            UNITER$_{large}$    & 303M  & 32.6  & 26.1  & 8.7   & 10.0  & -     & -     & -     & 49.1  & 17.9  & 0.0   & - \\
            VL-T5               & 224M  & 28.2  & 27.5  & 8.2   & 0.0   & 0.0   & 0.0   & 1.1   & 48.7  & -     & 13.5  & 4.4 \\
            % VL-T5               & 224M  & 28.2  & 27.5  & 8.2   & 0.0   & 0.0   & 0.0   & 0.0   & & -     & & 4.4 \\
            % VL-T5 (with prompt) & 224M  &       &       &       &       & 0.0   & 0.0   & 1.1   & & -     & & - \\
            FewVLM$_{base}$     & 224M  & 25.9  & 25.4  & 6.5   & 0.0   & 0.0   & 0.0   & 0.0   & 50.6  & -     & 43.4  & 31.0 \\
            FewVLM$_{large}$    & 740M  & 27.0  & 26.1  & 7.4   & 0.0   & 0.0   & 0.0   & 0.0   & 51.2  & -     & \bf 47.7  & \bf 36.5 \\
            % Oscar$_{large}$     & - & - & - & - & 49.4 & - & - & - & - & 0.0 & - \\
            % MDETR-ENB3          & - & - & - & 54.0$^\ddagger$ & - & 84.8$^\dagger$ & 93.8$^\dagger$ & 95.6$^\dagger$ & - & - & - \\ 
            % GLIP-L              & - & - & - & - & - & 87.1$^\dagger$ & 96.9$^\dagger$ & 98.1$^\dagger$ & - & - & - \\
            % Flamingo            & - & - & - & - & - & - & - & - & - & 56.3 & 67.2 \\ 
            \midrule
            \methodn   & 310M  & \bf 40.6  & \bf 39.3  & \bf 16.2  & \bf 47.5  & \bf 18.9  & \bf 53.4  & \bf 70.3  & \bf 56.1  & \bf 46.9  & 42.3  & 25.6 \\
            % \methodn$_{large}$  & 825M  & 39.7  & 39.6  & 16.3  & 47.0  & 16.8  & 50.4  & 68.2  & 57.0  & 46.9  & 49.3  & 26.6 \\
            \bottomrule
        \end{tabular}
	}
	\caption{\textbf{Zero-shot results.} We report performance on downstream tasks without any training data. Our model surpasses all baselines on classification tasks. 
 % * is from \cite{jin2021good}. $^\dagger$These models used the Flickr30k-entities dataset in the pre-training while ours did not. $^\ddagger$This model used the RefCOCOg dataset in the pre-training.
	}
	\label{tab:zero-shot}
\end{table*}

\subsection{Experiment Details}
% \wj{how to create hybrid seq}
For pre-training, we use 1,280 batch size for \method, set learning rate 1e-4 with 5\% linear warmup, and pre-train it with 30 epochs.
% and 960 batch size for \methodn$_{large}$ and pre-train them with 30 epochs.
% We set learning rate 1e-4 for the base model and 5e-5 for the large model with 5\% linear warmup in pre-training
For the few-shot setting, we randomly choose 32 examples and sample 5 different training and dev splits, and
we train models with 100 epochs with a learning rate of 5e-5 and choose the best checkpoint using the dev split.
\method has 310M parameters.
% and \methodn$_{large}$ has 825M parameters.
% For baselines, we use their official codes to get zero-shot and few-shot performance.
% \wj{number of fewshots}
% For pre-training, we set batch size 1,280 and 800 for \method$_{base}$ and \method$_{large}$, respectively and pre-train them with 30 epochs. 
% We use learning rate 1e-4 with 5\% linear warmup.
% For few-shot learning, we train models with 200 epochs, learning rate 5e-5 and 5\% linear warmup and choose the best checkpoint on the dev set.
% For \method, we use ``question: \texttt{[Q]} answer \texttt{<text\_1>}'' (P3) as an input prompt and ``\texttt{<text\_1>} \texttt{[A]}'' as a target prompt for visual question answering, and ``an image of'' (Q3) as an input prompt for captioning, which show the best performance. 
% We will study the effect of different prompts in Sec.~\ref{sec:exp:prompts}.
% The sizes of of $\mathcal{D}_{train}$ and $\mathcal{D}_{dev}$ are 16 on VQA and captioning tasks.
% For miniImageNet, we use `This is \texttt{<text\_1>},'' and ``\texttt{<text\_1>} \texttt{[A]}'' as input and target prompts.
% In this data, we test with \{1, 3, 5\}-shots per class. 

% \wj{parameter size of our model, 310M, 825M}
% \wj{implementation details}

\subsection{Evaluation Setup}
To evaluate few-shot performance, we randomly sample 5 different training and dev splits and measure the average performance on the 5 splits. 
We fine-tune the vision-language models with 100 epochs for the few-shot setup and choose the best checkpoint on the dev set.
% We take the average of five runs in the few-shot learning.
% For NoCaps task, it does not have training data. Thus we use the training data from COCO captioning in the experiments following~\citept{wang2021simvlm}. 
%We train the models with training data from COCO captioning for NoCaps following~\citept{wang2021simvlm} since NoCaps does not have its own training data.
We report the model performance on the test set for RefCOCOg, NLVR2, Flickr30k-entities, SNLI-VE, and Flickr30k captioning (Karpathy split~\citep{karpathy2015deep}), and the validation set for VCR and VQAv2.
% We evaluate on VCR validation set, RefCOCOg test set, NLVR2 test set, Flickr30k-entities test set, SNLI-VE test set,  VQAv2 validation set, and test set of Karpathy split for Flickr30k captioning.
% We use BLEU-4~\citep{papineni2002bleu}, CIDEr~\citep{vedantam2015cider}, METEOR~\citep{banerjee2005meteor}, SPICE~\citep{anderson2016spice} as evaluation metrics for captioning.
We adopt accuracy for VCR, RefCOCOg, SNLI-VE, NLVR2, and VQA datasets; Recall$@$1,5,10 for Flickr30k-entities; and CIDEr~\citep{vedantam2015cider} for captioning as evaluation metrics.

\subsection{Baselines}
For baselines, we include existing VL models: UNITER$_{large}$~\citep{chen2019uniter}, VL-T5~\citep{cho2021unifying}, 
% Oscar$_{large}$~\citep{li2020oscar}, 
GLIP-L~\citep{li2022grounded,zhang2022glipv2}, MDETR-ENB3~\citep{kamath2021mdetr}; and few-shot VL models: FewVLM~\citep{jin2021good}, Flamingo~\citep{alayrac2022flamingo}, and CPT~\citep{yao2021cpt}.
% For a fair comparison, we pre-train VL-T5 using their code
For a fair comparison, we exclude VQA datasets for VL-T5 and pre-train the model using their code.
Parameter sizes of each model are 303M for UNITER$_{large}$, 224M for VL-T5, 
% 340M for Oscar$_{large}$, 
231M for GLIP-L, 152M for MDETR, 224M and 740M for FewVLM$_{base}$ and FewVLM$_{large}$, 3B and 80B for Flamingo, and 113M for CPT.
% \wj{113M CPT~\citep{yao2021cpt}}
% To our knowledge, there is no few-shot models for grounding tasks, so 
% UNITER~\citep{chen2019uniter}, VL-T5~\citep{cho2021unifying}, FewVLM~\citep{jin2021good}, Oscar+~\citep{zhang2021vinvl}, GLIP~\citep{li2022grounded,zhang2022glipv2}, MDETER~\citep{kamath2021mdetr}, Flamingo~\citep{alayrac2022flamingo}

\begin{table*}[!t]
	\centering
	\small
	\resizebox{\textwidth}{!}{
		\begin{tabular}{l|c|ccccccccccc}
            \toprule
           \multirow{2}{*}{\textbf{Method} } & \multirow{2}{*}{\textbf{Size} } &  \multicolumn{3}{c}{\textbf{VCR}} & \multicolumn{1}{c}{\textbf{RefCOCOg}} &  \multicolumn{3}{c}{\textbf{Flickr30k-entities}}  & \multicolumn{1}{c}{\textbf{NLVR2}} & \multicolumn{1}{c}{\textbf{SNLI-VE}} & \multicolumn{1}{c}{\textbf{VQAv2}} & \multicolumn{1}{c}{\textbf{Flickr30k}}\\
            \cmidrule(lr){3-5} \cmidrule(lr){6-6}     \cmidrule(lr){7-9} \cmidrule(lr){10-10}  \cmidrule(lr){11-11} \cmidrule(lr){12-12} \cmidrule(lr){13-13} 
            & & Q $\rightarrow$ A & QA $\rightarrow$ R & Q $\rightarrow$ AR & Acc & R@1 & R@5 & R@10 & Acc & Acc & Acc & CIDEr \\
            \midrule
            Random              & -     & $25.0$  & $25.0$  & $6.3$   & $19.0$  & $6.5$   & $27.7$  & $47.8$  & $50.0$  & $33.3$  & $0.0$     & - \\
            UNITER$_{large}$    & 303M  & $29.1_{\pm3.4}$  & $28.6_{\pm2.0}$  & $8.4_{\pm1.0}$   & $45.4_{\pm4.0}$  & -     & -     & -     & $53.1_{\pm9.3}$  & $40.7_{\pm8.4}$  & $24.2_{\pm3.9}$    & - \\
            VL-T5               & 224M  & $29.7_{\pm1.3}$  & $28.0_{\pm1.6}$  & $8.7_{\pm0.8}$   & $\mathbf{56.9}_{\pm2.0}$& $\mathbf{28.1}_{\pm2.7}$ & $60.6_{\pm2.6}$& $73.3_{\pm1.8}$ & $48.7_{\pm0.1}$  & -     & $43.7_{\pm1.8}$    & $28.0_{\pm1.2}$ \\
            % VL-T5               & 224M  & 29.7  & 28.0  & 8.7   & 27.3  & 23.6  & 56.4  & 70.2  & -     & 27.9 \\
            % VL-T5 (with prompt) & 224M  &       &       &       &\bf 56.9&\bf 51.0 & 72.8& 78.6 & -     & - \\
            FewVLM$_{base}$     & 224M  & $29.1_{\pm0.9}$  & $28.4_{\pm1.1}$  & $8.5_{\pm0.4}$   & $16.0_{\pm3.7}$  & $4.2_{\pm1.2}$   & $18.7_{\pm1.8}$  & $31.7_{\pm2.0}$  & $50.3_{\pm0.7}$  & -     & $47.8_{\pm0.2}$    & $37.5_{\pm2.9}$ \\
            FewVLM$_{large}$    & 740M  & $30.0_{\pm2.7}$  & $30.1_{\pm2.5}$  & $9.3_{\pm1.5}$   & $17.4_{\pm1.1}$  & $5.1_{\pm1.1}$   & $22.7_{\pm4.0}$  & $38.0_{\pm5.8}$  & $51.3_{\pm1.2}$  & -     & $\mathbf{52.3}_{\pm0.8}$    & $\mathbf{38.4}_{\pm2.1}$ \\
            % Oscar$_{large}$     & -     & -     & -     & -     & -     & -     & -     & -     & - \\
            % MDETR-ENB3          & - & - & - & 75.3 & - & 84.8$^\dagger$ & 93.8$^\dagger$ & 95.6$^\dagger$ & - & - & - \\
            % GLIP-L              & - & - & - & - & - & 87.1$^\dagger$ & 96.9$^\dagger$ & 98.1$^\dagger$ & - & - & - \\
            % Flamingo            & - & - & - & - & - & - & - & - & - & 67.6 & 75.4 \\ 
            \midrule
            \methodn   & 310M  & $\mathbf{41.1}_{\pm0.7}$  & $\mathbf{40.4}_{\pm1.1}$  & $\mathbf{16.7}_{\pm0.6}$  & $48.1_{\pm1.2}$  & $25.4_{\pm1.0}$  & $\mathbf{61.3}_{\pm1.8}$  & $\mathbf{76.0}_{\pm1.5}$  & $\mathbf{56.2}_{\pm0.3}$  & $\mathbf{48.4}_{\pm1.0}$  & $46.8_{\pm0.1}$    & $37.1_{\pm1.5}$ \\
            % \methodn$_{large}$  & 825M  & 41.1  & 40.7  & 17.3  & 46.6  & 25.6  & 57.5  & 71.8  & 56.7  & 49.9  & 49.1    & 37.5 \\
            \bottomrule
        \end{tabular}
	}
	\caption{\textbf{Few-shot results.} We report performance on downstream tasks with 32 labeled examples for fine-tuning. 
 % $^\dagger$We report zero-shot results since their model is already trained on the Flickr30k-entities. Flamingo used 16 examples in this setup. 
	}
	\label{tab:32shot}
\end{table*}

\subsection{Downstream Tasks and Datasets}

In this section, we compare our \method on 7 downstream tasks; visual Commonsense Reasoning, referring expression comprehension, phrase grounding, NLVR2, SNLI-VE, VQA, and captioning.
% VQA and captioning require generation for our method, while other datasets are classification tasks.

\para{Visual Commonsense Reasoning (VCR).}
Visual Commonsense Reasoning (VCR)~\citep{zellers2019recognition} is a multiple-choice question-answering task that requires commonsense reasoning between objects in images.
The task is decomposed into two sub-tasks, question answering (Q $\rightarrow$ A) and rationale prediction (QA $\rightarrow$ R). 
In the holistic setting (Q $\rightarrow$ AR), models have to predict answers and rationales. 
Following VL-T5~\citep{cho2021unifying}, we rank the choices with $P(\text{true})/(P(\text{true})+P(\text{false}))$. and choose the one with the highest score.
VCR provides bounding boxes around entities, with explicit groundings between those entities and references in questions.
% \wj{task format?}
% We evaluate modes on the validation set.
% \wj{Val set}
% \wj{Writing for exp results will be here}

\para{Referring Expression Comprehension.}
Referring expression comprehension is to localize an object given a referring expression. 
We adopt the RefCOCOg dataset~\citep{mao2016generation} for this task.
We present a referring phrase and candidate regions from the image to our model; our model finds the most plausible region to the given phrase by ranking the regions with $P(\text{true})/(P(\text{true})+P(\text{false}))$.
Following VL-T5~\citep{cho2021unifying}, we use Mask R-CNN~\cite{anderson2018bottom} to find region detections as candidates for inference.
We consider the selected region to be correct if its intersection over union (IoU) with the ground truth region is greater than 0.5.
The upper bound performance on the test set by the Mask R-CNN is 86.09\%.
We get the performance of the random predictor by randomly choosing the bounding box from the object detector.
% \wj{testset}
% \wj{val oracle: 85.95, test oracle: 86.09}
% \wj{random number is with mask R-CNN}
% \wj{MDETR included annotations from RefCOCO in pre-training}
% \wj{Writing for exp results will be here}

\para{Phrase Grounding.}
Given one or more phrases, phrase grounding is to provide a set of bounding boxes for each phrase.
We use the Flickr30k-entities dataset~\citep{plummer2015flickr30k} for this task.
Following BAN~\citep{kim2018bilinear} and VisualBERT~\citep{li2019visualbert}, we adopt Faster R-CNN~\citep{ren2015faster} pre-trained on Visual Genome to detect regions as candidates for inference.
The predicted region is correct if its intersection over union (IoU) with the ground-truth region is greater than 0.5. 
The upper bound performance on the test set by the Faster R-CNN is 87.45\%.
Similar to RefCOCOg we provide a referring phrase and candidate regions from the image to our model; and our model finds the most plausible region to the given phrase by ranking the regions with $P(\text{true})/(P(\text{true})+P(\text{false}))$.
We use the any-box-protocol from MDETR~\citep{kamath2021mdetr}.
% \wj{choose 32 sentences - it might have more pairs}
% \wj{Maybe merged box version?}
% \wj{upper bound: 87.45}
% \wj{mdeter zero-shot}

% \subsection{Object detection} 
\para{NLVR2.}
The task of NLVR2~\citep{suhr2018corpus} is to determine whether a text description is true given two images.
The task requires understanding two images and comparing them.
To apply our model to this task, we create one image by concatenating the two images, and then our model generates text labels ``true'' and ``false'' for inference.
% To apply our model to this task, we create one image by concatenating the two images and replacing the ``left'' word with the left image and the ``right" word with the right image. \wj{more}
% Then our model generates text labels ``true'' and ``false'' for inference.
% \wj{Writing for exp results will be here}

\para{Visual Entailment.}
Visual entailment, SNLI-VE~\citep{xie2019visual}, is to determine whether the image semantically entails the text given an image-sentence pair. 
The task is a 3-way classification where labels are ``entailment'', ``neutral'', and ``contradiction.''
We define label words for the classification as ``entailment'': ``true'', ``neutral'': ``maybe'', ``contradiction'': ``false.''
We choose the classification label by measuring the probability of each word and picking the highest one.

% \wj{Writing for exp results will be here}

\para{Visual Question Answering.}
The visual question answering task~\citep{goyal2017making} requires models to answer a question to a given context image.
% Recent approaches~\citep{chen2019uniter,tan2019lxmert,su2019vl,li2019visualbert,li2020oscar} tackle visual question answering tasks as multi-label classification over a predefined set of answer candidates. 
% Instead, 
We approach the visual question answering task as a generation task so that the model can produce the answers without introducing any task-specific heads following \citet{jin2021good,cho2021unifying}.
We adopt the input prompt, ``\textit{question:} \{question\} \textit{answer:} \texttt{<text\_1>},''  where \texttt{<text\_1>} is a sentinel token, from \cite{jin2021good} for the generation.
% ; this lets the model quickly learn given only a few examples.
% In this setup, prompts act as constraints to guide the models to generate proper formats of answers;  models might generate a sentence for VQA, which is not the correct format, without prompts.  

\para{Captioning.}
The captioning task is to generate a caption given an image.
In Flickr30k~\citep{young2014image}, 
we use ``\textit{an image of}' as our input prompt from \citet{jin2021good}.
% we explore three hand-crafted input prompts: ``\textit{a picture of}'', ``\textit{a photo of}'', and ``\textit{an image of}''. 
% \xiang{here what exactly happen is vague and that's why I ask for a Sec 4.4 to give the context first.}
% We study the effect of different word choices in this captioning task. While the three different words have similar meanings, they show different performance in zero-shot and few-shot tasks as we will see in our experiments..
% For target prompts, we just train the model with the original caption without any additional prompts. 

\begin{table}[tb!]
\centering
\resizebox{\linewidth}{!}{
\begin{tabular}[t]{l|c|cccc}
\toprule
    \multirow{2}{*}{\textbf{Method} }  & \multirow{2}{*}{\textbf{Size} } & \multicolumn{2}{c}{\textbf{RefCOCOg}} &\multicolumn{2}{c}{\textbf{Flickr30k-entities}}     \\
    \cmidrule(lr){3-4} \cmidrule(lr){5-6}
    & & 0 & 32 & 0 & 32  \\
\midrule
Random                                          & - & 19.0  & 19.0  & 6.5   & 6.5  \\
UNITER$_{large}$~\citep{chen2019uniter}         & 303M  & 10.0  & 45.4  & -     & -  \\
VL-T5~\citep{cho2021unifying}                   & 224M  & 0.0   & \bf 56.9  & 0.0   & \bf 28.1 \\
% FewVLM$_{base}$     & 43.4    & 47.8   \\
FewVLM$_{large}$~\citep{jin2021good}            & 740M  & 0.0   & 17.4  & 0.0   & 5.1\\
% Oscar$_{large}$     & 0.0   & 24.5  \\
CPT~\citep{yao2021cpt}~\citep{yao2021cpt}       & 113M  & 36.5  & -     & -     & - \\
MDETR-ENB3~\citep{kamath2021mdetr}              & 152M  & \bf 54.0$^\dagger$ & -     & 84.8$^\ddagger$ & - \\
GLIP-L~\citep{li2022grounded,zhang2022glipv2}   & 231M  & -     & -     & \bf 87.1$^\ddagger$ & - \\
% Flamingo \\
\midrule
\methodn                                        & 310M  & 47.5  & 48.1  & 18.9  & 25.4  \\
% \methodn$_{large}$                              &   &  \\
\bottomrule
\end{tabular}}
\caption{\textbf{Results on RefCOCOg and Flickr30k-entities with 0 and 32 examples.} We report recall@1 for Flickr30k-entities. $^\dagger$This model used the RefCOCOg dataset in the pre-training. $^\ddagger$These models used the Flickr30k-entities dataset in the pre-training while ours did not.  
% \wj{run 32shot refcoco again}
}
\label{tab:groundingtask}
\end{table}

\begin{table}[tb!]
\centering
\resizebox{0.9\linewidth}{!}{
% \begin{tabular}[t]{L{3cm}|C{1.5cm}C{1.5cm}C{1.5cm}}
\begin{tabular}[t]{l|c|ccc}
\toprule
    \textbf{Model}   & \textbf{size} &\textbf{0-shot}    &\textbf{32-shot }      \\
\midrule
Random                                  & -     & 0.0   & 0.0  \\
UNITER$_{large}$~\citep{chen2019uniter} & 303M  & 0.0   & 24.2  \\
VL-T5~\citep{cho2021unifying}           & 224M  & 13.5  & 43.7 \\
% FewVLM$_{base}$     & 43.4    & 47.8   \\
FewVLM$_{large}$~\citep{jin2021good}    & 740M  & 47.7  & 52.3 \\
% Oscar$_{large}$~\citep{li2020oscar}     & 0.0   & 24.5  \\
Flamingo-3B~\citep{alayrac2022flamingo}    & 3B &49.2  & 57.1 \\
Flamingo-80B                            & 80B   &\bf 56.3  & \bf 67.6\\
\midrule
\methodn                                & 310M  & 42.3  & 46.8 \\
% \methodn$_{large}$                      & 49.3  & 49.1 \\
\bottomrule
\end{tabular}}
\caption{\textbf{VQA results with 0 and 32 examples.} We report zero-/32-shot performance on the VQAv2 dataset. Flamingo has 3B or 80B parameters and uses in-context examples for inference while our model has 310M parameters and uses the examples for fine-tuning. 
    % \xiang{1. probably just focus on large models and remove others; 
    % 2. do we have quoted numbers for other baselines here? 
    % 3. Add citations in this table too.}
}
\label{tab:vqa}
\end{table}

\subsection{Results}

\para{Zero-shot performance.}
We evaluate the existing models in a zero-shot manner, where models do not have access to any training data.
Tab.~\ref{tab:zero-shot} shows the performance on each task.
% Note that VCR, RefCOCOg, NLVR2, Flickr30k-entities require phrase or region grounding.
First, \method shows the best performance on most tasks while baselines show worse performance than the random predictor on many of the grounding tasks.
On Table~\ref{tab:groundingtask}, we additionally include baselines, GLIP-L and MDETR-ENB3, that are targeted for grounding tasks.
These models include the corresponding task-specific datasets in pre-training so they demonstrate great performance without additional fine-tuning.
Note that we do not include task-specific datasets in the pre-training.
% Our method also exhibits decent performance on VQAv2 and Flickr30k captioning. 
In addition, our model still performs well on SNLI-VE, visual question answering and captioning that do not require explicit grounding.
% shows a better performance on VCR compared to other baselines.
By comparing Flamingo in Tab.~\ref{tab:vqa}, a 3B or 80B-sized vision-language model, our model demonstrates good accuracy considering our model size.
% On Flickr30 captioning, our model underperforms FewVLM$_{base}$ which is a bit smaller model than ours.
This suggests that our model has a generalization capability to unseen tasks while competitors have difficulty generalizing to grounding tasks that need phrase or region grounding in a zero-shot way.
% We 

% We include Flamingo, 80B vision-languge model, on Table~\ref{tab:vqa}, 

% VQAv2, flickr30k captioning
% Flamingo

% visual commonsense reasoning which needs to classify correct referring words and each answer option.
% We also observe the better performance on NLVR2. 
% The NLVR2 task asks about two images and classify whether the given sentence is correct.
% MDETR-ENB3 shows good performance on RefCOCOg and Flickr30k-entities and GLIP-L also exhibits good performance on Flickr30k-entities.
% These models include the corresponding task-specific datasets in pre-training so they demonstrate great performance without additional fine-tuning.
% Note that we do not include the task-specific datasets in the pre-training.
% On VQA, our method also shows comparable results to Flamingo which has 80B parameters, 97 times larger than our model.
% Our model sacrifices the performance on Flickr30k captioniong.
% Comparing our base and large models, we observe similar results on the zero-shot tasks except for the VQA task.
% \wj{mention glip, mdetr, flamingo}
% \wj{table~\ref{tab:vqa},  \ref{tab:groundingtask}}

\para{Few-shot performance.}
We evaluate our model and competitors on the few-shot setting (Tab.~\ref{tab:32shot}).
% We use 32 labeled examples in total for fine-tuning.
Our model, \methodn, shows great performance overall, while VL-T5 outperforms our model on the RefCOCOg dataset
% While our model, \methodn, improves the performance on all the tasks, baseline methods outperforms our model on RefCOCOg and Flickr30k-entities unlike zero-shot results.
We conjecture that the method includes the phrase grounding task in their pre-training, so it achieves great performance.
However, the model still struggles with other tasks including the VCR task, which demonstrates their limited generalization.
% while our model surpasses the models on the task.
Our model shows consistently good results and thus exhibits great generalization on the few-shot setup.
% Interestingly, our model achieves the comparable result to FewVLM on the Flickr30k captioning on the few-shot setup. 
% Thes few-shot

% Table~\ref{tab:32shot} shows the few-shot performance. We use 32 examples in total for fine-tuning.
% Overall, we observe similar trend as in the zero-shot setup. 
% On the few-shot setting, our model improves particularly on Flickr30k-entities, VQA, Flickr30k captioning tasks compared to zero-shot performance.
% Furthermore, our model outperforms other baselines.
% Interestingly, our large model rather underperforms the base model on Flickr30k-entities and it shows similar performance on other datasets except VQA.

% \wj{flamingo 3b captionnig: 60.6, 71.2. 80b 67.2, 75.4}

\subsection{Ablations}
\begin{figure}[tb!]
    \centering
    {\includegraphics[width=0.7\linewidth]{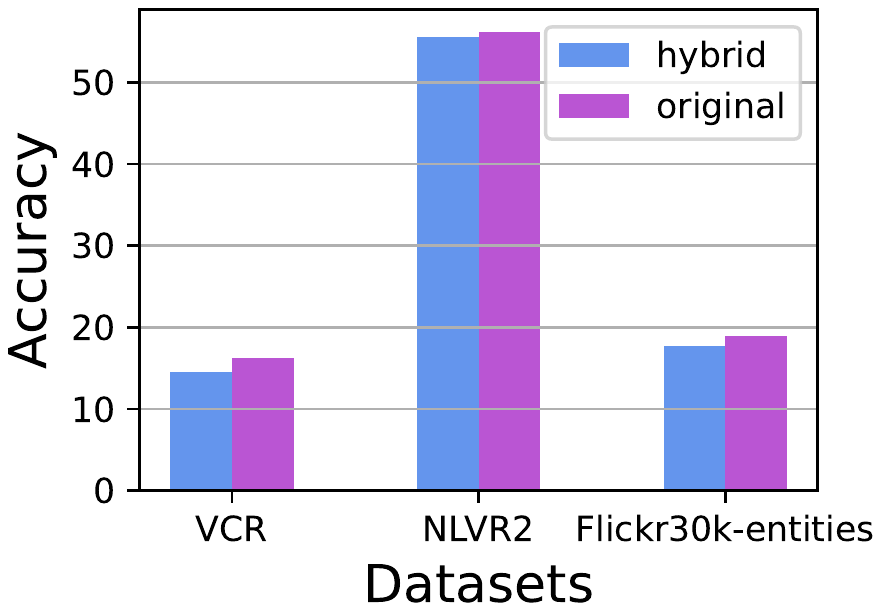}} 
\caption{\textbf{Performance with different input formats for inference on the zero-shot setup.}  We report Q $\rightarrow$ AR for VCR, and R@1 for Flick30k-entities. }
    \label{fig:nogrounding}
\end{figure}

Here, we study ablations for our method.
Tab.~\ref{tab:nodiscground} and Fig.~\ref{fig:nogrounding} show the ablations on the hybrid sequences and pre-training objectives, and different input formats during inference on the zero-shot setup, respectively.

\para{Hybrid sequences and pre-training objectives.}
We study the ablation of pre-training objectives and hybrid sequences in pre-training.
On Tab.~\ref{tab:nodiscground}, our model without hybrid sequences significantly affects the performance on many tasks.
Specifically, results on RefCOCOg and Flickr30k-entities are significantly degraded, suggesting that hybrid sequences in pre-training play a vital role in improving phrase grounding.
Among pre-training objectives in \method, we notice that the discriminative objective is important for many of the tasks while others do not affect a lot.
We conjecture that the tasks in the table are classification tasks so the discriminative objective is the most useful for the tasks.

% hybrid sequences in pre-training data which include object bounding boxes are important for RefCOCOg and Flickr30k-entities. 

% As shown in the table, the discriminative objective is important for all the tasks, especially, VCR and NLVR2.
% Without this objective, the performance on VCR and NLVR2 is significantly degraded, which suggest that classification-type of pre-training is required for these tasks.
% On the other hand, grounded sequences in pre-training data which include object bounding boxes are important for RefCOCOg and Flickr30k-entities. 
% We conjecture that the RefCOCOg and flickr30k-entities tasks are to find correct bounding boxes and corresponding descriptions, so learning such information in pre-training is vital for the tasks.
% Interestingly, VCR and NLVR2 do not require such pre-training on bounding boxes for the better performance.

% \wj{negative sampling}
% \wj{no itm}
% \wj{no grounded seq}

% \subsection{Analysis}
\para{Input formats in inference.}
We investigate the different input formats (hybrid sequences vs. original sequences) during zero-shot inference on Fig.~\ref{fig:nogrounding}.
Note that we use hybrid sequences in the pre-training.
% Here we discuss the results on the input formats (Fig.~\ref{fig:nogrounding}.
On VCR, we replace the referring words (e.g., [person1] in Fig.~\ref{fig:intro_example})  with bounding boxes for text input (hybrid sequences), or we do not replace them and use original text input (original sequences).
On NLVR2, we replace the ``left'' word with the left image and the ``right" word with the right image (hybrid sequences), or we do not replace them and use the original text input (original).
On Flickr30k-entities, we replace the referring words with corresponding bounding boxes (hybrid sequences), or we don't replace the referring words and use the referring words and bounding boxes for inference (original).
Counter-intuitively, we observe that our model with original input formats during inference shows better performance on all the datasets.
We conjecture that using the hybrid sequences with bounding boxes may disturb the model predictions since the model needs to judge whether the grounding information is correct or not.
% information is correct or not.
We leave the sophisticated design for future work.

\begin{table}[tb!]
\centering
\resizebox{1\linewidth}{!}{
\begin{tabular}[t]{L{4cm}|C{1.5cm}C{1.5cm}C{1.5cm}C{1.5cm}}
\toprule
    \textbf{Model}   &\textbf{VCR }   &\textbf{ RefCOCOg }    &\textbf{NLVR2 }  &\textbf{Flickr30k-entities }     \\
\midrule
\textbf{Zero-shot} \\
\methodn                & 16.2  & 47.5 & 56.1 & 18.9    \\
No hybrid sequences   & 12.9  & 18.9 & 55.7 & 5.7   \\
No discriminative       & 6.8   & 30.5 & 50.4 & 12.7    \\
No PrefixLM             & 14.4  & 48.5 & 55.8 & 18.5 \\
No MLM                  & 15.6  & 47.8 & 56.0 & 19.3\\
\midrule
\textbf{32-shot} \\
\methodn                & 16.7  & 48.1 & 56.2 & 25.4    \\
No hybrid sequences   & 14.3  & 16.3 & 55.9 & 18.7  \\
No discriminative       & 7.2   & 42.0 & 50.5 & 15.3   \\
No PrefixLM             & 14.7  & 48.7 & 55.9 & 21.9\\
No MLM                  & 16.3  & 47.9 & 56.1 & 23.5 \\
\bottomrule
\end{tabular}}
\caption{\textbf{Ablations on the pre-training objectives and hybrid sequences in pre-training.} We report Q $\rightarrow$ AR for VCR, and R@1 for Flick30k-entities.} 
\label{tab:nodiscground}
\end{table}

% \begin{table}[tb!]
% \centering
% \resizebox{0.8\linewidth}{!}{
% \begin{tabular}[t]{L{3cm}|C{1.5cm}C{1.5cm}C{1.5cm}C{1.5cm}}
% \toprule
%     \textbf{Model}   &\textbf{VCR }    &\textbf{NLVR2 }  & \textbf{Flickr30k-entities}    \\
% \midrule
% Hybrid      & 14.5 & 55.5 & 17.7    \\
% Original    & 16.2 & 56.1 & 18.9    \\
% \bottomrule
% \end{tabular}}
% \caption{\textbf{Performance with different input formats for inference on the zero-shot setup.}  We report Q $\rightarrow$ AR for VCR, and R@1 for Flick30k-entities. \wj{32, and 64 shot}}
% \label{tab:nogrounding}
% \end{table}

\section{Related Work}

% Flamingo~\citep{alayrac2022flamingo}
% GLIP~\citep{li2022grounded,zhang2022glipv2}

% Comparison to FewVLM
% \wj{difference from glip, oscar, etc}

% Why not object detection

% the localization loss Ll trains localization heads with bounding-box supervision, e.g., RPN loss, box regression loss and/or centerness loss

\para{Vision-language few-shot learning.}
There have been attempts to address the challenge of data-hungry supervised learning in vision-language domains: FewVLM~\cite{jin2021good}, Frozen~\cite{tsimpoukelli2021multimodal},  Flamingo~\citep{alayrac2022flamingo}, GLIP~\citep{li2022grounded,zhang2022glipv2},
FewVLM~\citep{jin2021good} improves the few-shot performance of VQA and captioning by prompting the model and its performance is on par with large few-shot learners.
Frozen~\citep{tsimpoukelli2021multimodal} adapts a few-shot language model~\citep{radford2019language} to vision-language tasks with soft prompting for images.
% Their approach shows the few-shot capability on visual question answering and image classification tasks.
% Similarly, PICa~\cite{yang2021empirical} uses GPT-3~\cite{brown2020language} to solve VQA tasks in a few-shot manner by providing a few in-context VQA examples. 
% It converts images into textual descriptions so that GPT-3 can understand the images.
% SimVLM~\citep{wang2021simvlm} is trained with prefix language modeling on weakly-supervised datasets and 
% it exhibit surprising results on a zero-shot captioning task.
Flamingo~\citep{alayrac2022flamingo} achieves state-of-the-art results on few-shot VQA and captioning tasks by prompting the model with task-specific examples.
While these models achieve improvement on few-shot tasks, they are not applicable to grounding tasks.
% are impractical to use in real-world applications due to their model sizes.
Lastly, GLIP~\citep{li2022grounded,zhang2022glipv2} unifies object detection and phrase grounding and it achieves great performance on zero-shot object detection and phrase grounding tasks. 
Unlike our method, GLIP used grounding datasets including Flickr30k-entities in pre-training so it achieved great performance on the phrase grounding without fine-tuning.
Our method is not applicable to object detection since it requires bounding box regression. 
We leave this extension for future work.

\para{Grounded vision-language learning.}
Grounded vision-language learning has been explored to learn grounding between objects in images and phrases in sentence~\citep{li2020oscar,zhang2021vinvl,kamath2021mdetr,li2022grounded,zhang2022glipv2}.
% Oscar use object tags and region features as additional context to align the image and language modalities in a shared semantic space~\citep{li2020oscar,zhang2021vinvl}.
% It achieves great performance on visual question answering and captioning but it is not generalized to grounding tasks.
MDETR is a modulated detector that detects objects in an image conditioned on a raw text query~\citep{kamath2021mdetr}.
The model exhibits remarkable results on object detection, phrase grounding, and referring expression comprehension by pre-training the model on object detection data. 
GLIP followed a similar direction and it unifies object detection and phrase grounding~\citep{li2022grounded,zhang2022glipv2}.
While the methods rely on object detection datasets to improve grounding, our method utilizes grounded sequences from image-caption datasets and an object. 
Our model does not only work on grounding tasks but also on visual question answering and captioning tasks.
% does not include object detection dataset and 

% \para{Pre-training vision-language models.} There is a tremendous amount of work in training generic models for a variety of vision-language tasks, such as visual question answering (VQA), and image captioning, etc. 
% Most of the existing methods employ BERT-like architectures \cite{devlin2018bert} to learn cross-modal representations from a concatenated sequence of visual region features and language token embeddings. 
% For example, early efforts such as VisualBERT, VL-BERT and Oscar \cite{su2019vl,li2019visualbert,li2020oscar} propose either a single-stream or two-stream Transformer-based framework. 
% \citet{chen2019uniter} conduct comprehensive
% studies on the effects of different pre-training objectives on the learned representations. 
% \citet{Cao2020BehindTS} design a set of meticulously designed probing tasks to decipher the inner workings of multimodal pre-training.  
% \citet{gan2020large} propose an adversarial learning framework to improve the vision-and-language representation. Instead of using the regional features extracted by pre-trained object detection models like Faster-RCNN, 
% SOHO \cite{huang2021seeing} proposes to jointly learn Convolutional Neural Network (CNN) and Transformer for cross-modal alignments from image-text pairs. 
% The UNIMO architecture \cite{li2020unimo} can leverage the large scale of non-paired text corpus and image collections for cross-modal learning.

% \section{Limitations}

% Why not object detection

\section{Conclusion}
In this work, we proposed \methodn, a new VL model that can  generalize to a variety of VL tasks including grounding tasks.
Our model learns object grounding and localization by introducing hybrid sequences in pre-training and easily adapt to diverse task by using a vision transformer for versatile image processing.
% Our model is a sequence-to-sequence transformer model that uses a vision transformer for versatile image processing on zero-shot tasks.
To pre-train our model, we introduced our dataset using object-word alignments and pre-train it with masked language modeling, prefix language modeling, and the discriminative objective.
On the empirical analysis, we observed that our model demonstrated good zero-/few-shot generalization on diverse tasks. 
We also observed that the discriminative objective and hybrid sequences in pre-training were vital for better zero-/few-shot performance.
% We leave the better design of input formats in inference for future work.
% \wj{include object detection}

\bibliographystyle{acl_natbib}
\bibliography{bibtex}

% \newpage
% \clearpage
% \appendix
% % \section{Appendix}
% \input{090appendix}
% \label{sec:appendix}

% This is an appendix.

\end{document}